\documentclass[10pt,twocolumn,letterpaper]{article}

\usepackage{cvpr}              

\usepackage{algorithm}
\usepackage{algorithmic}

\usepackage[utf8]{inputenc} 
\usepackage[T1]{fontenc}    
\usepackage{amsmath}
\usepackage{amsthm}

\usepackage{url}            
\usepackage{booktabs}       
\usepackage{amsfonts}       
\usepackage{nicefrac}       
\usepackage{microtype}      
\usepackage{xcolor}         
\usepackage[table]{xcolor}
\usepackage{graphicx}
\usepackage{diagbox}
\usepackage{url}
\usepackage{booktabs}
\usepackage{bbding}
\usepackage{pifont}
\usepackage{wasysym}
\usepackage{utfsym}
\usepackage{fontawesome}
\usepackage{appendix}
\usepackage{multirow}
\usepackage{enumitem}
\usepackage[breakable]{tcolorbox}
\usepackage{tabularx}
\usepackage{makecell}

\definecolor{mygreen}{RGB}{146, 199, 113} 
\definecolor{cvprblue}{rgb}{0.21,0.49,0.74}
\usepackage[pagebackref,breaklinks,colorlinks,allcolors=cvprblue]{hyperref}

\def\confName{CVPR}
\def\confYear{2027}

\title{Specificity-Guided Knowledge Diversion for Decomposable Generative Models}

\author{Yucheng Xie\textsuperscript{\rm 1,2}\;\; Fu Feng\textsuperscript{\rm 1,2}\;\; Jing Wang\textsuperscript{\rm 1,2}\thanks{Corresponding authors}\;\;\; Xin Geng\textsuperscript{\rm 1,2}\footnotemark[1]\;\;\; Yong Rui\textsuperscript{\rm 1,2}\\
\textsuperscript{\rm 1}School of Computer Science and Engineering, Southeast University, Nanjing, China\\
\textsuperscript{\rm 2}Key Laboratory of New Generation Artificial Intelligence Technology and Its Interdisciplinary \\Applications (Southeast University), Ministry of Education, China\\
{\tt\small \{xieyc, fufeng, wangjing91, xgeng\}@seu.edu.cn}\\
}

\begin{document}
\maketitle

\begin{abstract}
Generative model pre-training typically optimizes a model as a whole without explicitly organizing its parameters for selective reuse. 
Decomposable pre-training instead learns reusable components for downstream adaptation while retaining strong generative performance. 
However, existing knowledge diversion methods incur substantial parameter overhead through weight factorization and rely on fixed component assignments to induce specialization, potentially compromising in-distribution generation quality.
We therefore propose Specificity-Guided Knowledge Diversion (SGKDiv), which learns how generative knowledge is shared and specialized for effective generation and selective downstream reuse.
SGKDiv constructs Knowledge Units from paired feed-forward channels, preserving native computational structure without SVD-based reparameterization.
Calibrated condition-wise routing combines a shared usage prior with conditional adjustments, while complementary regularization shapes their allocation, diversity, and magnitude. 
Together, these designs replace predefined component assignments with adaptively learned and explicitly regulated knowledge specialization.
Experiments across diverse model and training settings demonstrate better in-distribution generation than dense and MoE baselines, alongside stronger downstream adaptation than representative PEFT methods across eight datasets.
\end{abstract} 

\vspace{-0.1in}
\section{Introduction}
Large-scale pre-training has become a prevailing paradigm for image generation, supporting both in-distribution generation and downstream adaptation~\cite{bu2026hiflow, xuan2026diff, xin2026parameter, qiu2020pre, han2021pre}. 
However, \textit{optimizing a model as a whole} does not explicitly distinguish transferable knowledge from source-specific information, limiting selective knowledge reuse~\cite{zarei2026localizing, ni2025makes}. 
This motivates \textit{decomposable model pre-training}~\cite{feng2026constraint, xie2026knowledge}, which organizes knowledge into modular components to retain shared capabilities while enabling selective adaptation to downstream requirements, including varying computational constraints~\cite{haji2026one, xie2025SANA} and target-domain shifts~\cite{sani26MMD, wang26adap}.

\begin{figure}[tb]
  \centering
  \includegraphics[width=\linewidth]{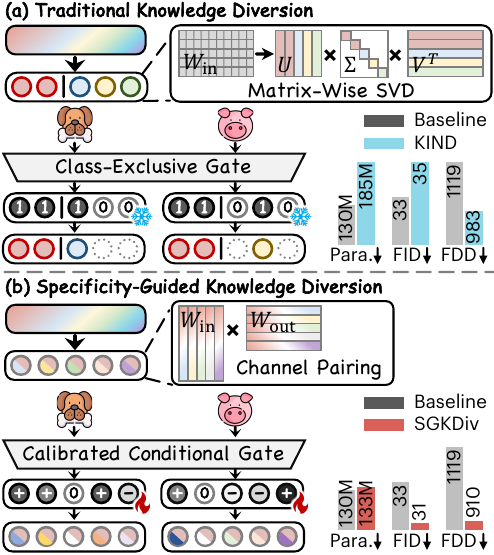}
  \vspace{-0.25in}
  \caption{
  (a)~Traditional knowledge diversion constructs SVD-based units with predefined shared--specific assignments, improving downstream adaptation (FDD) at the cost of degraded in-distribution generation (FID) and substantial parameter overhead.
  (b)~SGKDiv constructs units from paired native FFN channels and learns their contributions through calibrated routing, improving both generation and adaptation with modest parameter overhead.}
  \label{fig:moti}
  \vspace{-0.12in}
\end{figure}

These downstream requirements motivate complementary forms of decomposability. Constraint-based pre-training~\cite{feng2026constraint,feng2024wave,xie2026self,xie2026fine} addresses computational constraints through size-wise knowledge reuse, encoding structured weight templates for initialization across model depths and widths.
In parallel, Knowledge Diversion (KDiv)~\cite{xie2025kind,xie2026knowledge} enables condition-wise reuse by organizing generative knowledge into condition-agnostic and condition-specific components, preserving shared capabilities while adapting specialized components to target conditions and domains.

Despite supporting selective downstream reuse, existing knowledge diversion methods face two limitations (Fig.~\ref{fig:moti}a). 
First, matrix-wise SVD introduces redundant factor parameters and treats consecutive projections independently, overlooking their computational correspondence~\cite{xie2026knowledge, xie2026divcontrol,feng2025knowledge}.
Meanwhile, assigning knowledge units to fixed shared and condition-specific subsets before training restricts knowledge allocation and relies on structural separation alone to induce specialization~\cite{xie2025kind, shi2025fad, xu2025ctrlora}. 
Without explicit guidance on how condition-specific knowledge should be learned and utilized, this rigid organization can hinder generative learning and compromise in-distribution generation quality.

These limitations motivate \textbf{\textit{Specificity-Guided Knowledge Diversion (SGKDiv)}} (Fig.~\ref{fig:moti}b), which exploits native computational structure and explicitly regulates specialization.  
Through adaptive routing and regularization, SGKDiv allows shared usage and condition-specific specialization of Knowledge Units to emerge during training, without a predefined partition. 
This makes knowledge diversion a flexible pre-training approach with learnable, explicitly regulated knowledge allocation that supports downstream reuse while maintaining strong in-distribution generative performance.

Specifically, SGKDiv constructs \textit{Knowledge Units} by pairing each FFN input-projection column with its corresponding output-projection row, preserving native nonlinear computation without SVD-based reparameterization. 
Unlike token-wise mixture-of-experts (MoE)~\cite{wei2026routing, sun2025ecdit, fei2024scaling}, SGKDiv modulates Knowledge Units at the condition level.
To this end, a calibrated \textit{condition-agnostic gate} defines a shared usage prior, while a \textit{condition-specific gate} adjusts it in logit space to determine each unit's contribution.
To encourage effective specialization, we introduce gate polarization to sharpen the shared usage prior, together with deviation diversity and magnitude regularization to shape distinct conditional adjustments of appropriate strength. 
Such decomposable structure supports target-aware selection of Knowledge Units for parameter-efficient downstream fine-tuning.

We evaluate SGKDiv on class-conditional image generation across diverse model and training configurations, including DiT (-Flow), REPA, and JiT. 
SGKDiv achieves higher in-distribution generation quality than dense and MoE baselines with fewer activated parameters, demonstrating that decomposability can enhance rather than compromise generative learning.
For downstream adaptation, fine-tuning target-relevant Knowledge Units outperforms representative PEFT methods under comparable trainable parameter budgets, supporting the effectiveness of selective component reuse. 
Together, these results demonstrate that SGKDiv supports effective generative pre-training and parameter-efficient downstream adaptation within a unified framework.

Our main contributions are as follows:
1)~We propose \textit{Specificity-Guided Knowledge Diversion}, a generative pre-training framework that moves beyond predefined parameter allocation to learn reusable knowledge organization for both in-distribution generation and downstream adaptation.
2)~We introduce native paired-channel Knowledge Units that preserve FFN pathways without SVD-based reparameterization, together with calibrated routing that explicitly regulates shared allocation and conditional specialization.
3)~Experiments demonstrate improved generation quality over dense and MoE baselines, alongside stronger downstream adaptation than PEFT methods under comparable trainable parameter budgets.

\section{Related Work}

\paragraph{Conventional Pre-Training and Parameter-Efficient Adaptation.}
Large-scale pre-training enables vision models to learn transferable representations from labeled data~\cite{dosovitskiy2021an} or self-supervision through instance discrimination~\cite{chen2020simple}, masked prediction~\cite{bao2022beit,he2022masked}, and self-distillation~\cite{oquab2024dinov}.
In image generation, latent diffusion models~\cite{rombach2022high} similarly learn generative priors that support subsequent adaptation for personalization~\cite{ruiz2023dreambooth} and spatial control~\cite{zhang2023adding}. 
Such adaptation can be costly when the entire model is updated~\cite{touvron2021training,achiam2023gpt}. 
Parameter-efficient fine-tuning (PEFT) reduces this cost by restricting updates to prompts~\cite{jia2022visual}, adapters~\cite{houlsby2019parameter,chen2022adaptformer}, low-rank factors~\cite{hulora,hu2023llm}, or selected spectral components~\cite{han2023svdiff,robb2020few,sun2022singular,zhang2024spectrum,zhang2024spectral}.
Although these methods reduce the trainable parameter budget, they operate on conventionally pre-trained backbones without an explicit separation of transferable and source-specific knowledge, leaving limited downstream updates to accommodate target-specific requirements while mitigating source-domain bias.

\vspace{-0.1in}
\paragraph{Scalable Model Pre-training.}
Decomposable pre-training enhances the modularity of pre-trained models by learning reusable components.
The \textit{Learngene} framework~\cite{feng2023genes,feng2024transferring} encapsulates transferable knowledge in network fragments to support efficient downstream adaptation.
Within this framework, constraint-based pre-training~\cite{feng2026constraint} enables \textbf{\textit{cross-scale knowledge reuse}} by imposing structural constraints during training through linear transformations~\cite{xia2024transformer}, Kronecker-product factorizations~\cite{feng2024wave}, and other structured decompositions~\cite{xie2026fine,xie2026self,shen2026breaking,shen2026unified}.
Knowledge diversion (KDiv)~\cite{xie2026knowledge} pursues \textbf{\textit{cross-task knowledge reuse}} by separating condition-agnostic and condition-dependent knowledge into reusable components. 
KIND~\cite{xie2025kind} introduced this principle for class-conditional generation, while FAD~\cite{shi2025fad} and DivMorph~\cite{feng2025knowledge} extended it to frequency-aware adaptation and cross-morphology transfer, respectively. 
Our SGKDiv advances knowledge diversion through native paired-channel construction and specificity-regularized optimization, jointly guiding knowledge organization and specialization.

\section{Method}
\subsection{Preliminaries}

\begin{figure*}[tb]
\centering
\includegraphics[width=\linewidth]{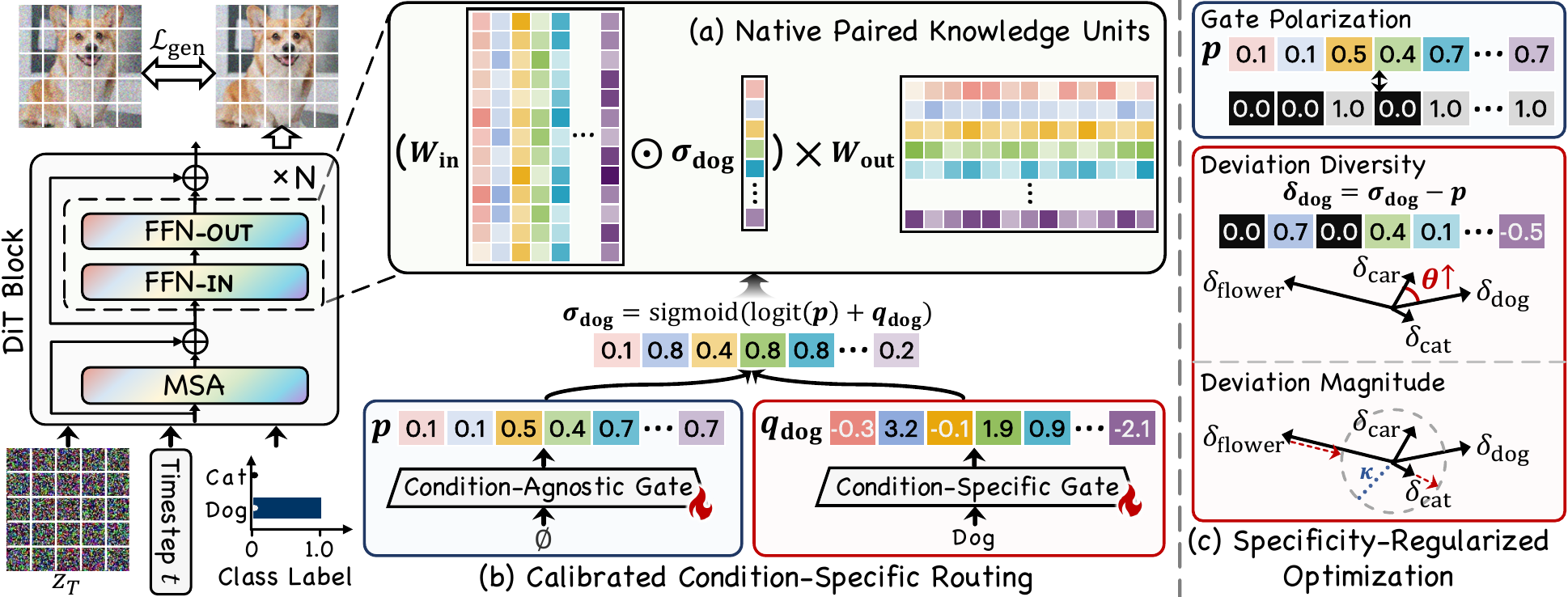}
\vspace{-0.2in}
\caption{\textbf{Overview of Specificity-Guided Knowledge Diversion (SGKDiv).}
(a)~Paired FFN input columns and output rows form native Knowledge Units without matrix factorization.
(b)~Calibrated condition-specific routing combines a shared prior $\boldsymbol{p}$ with conditional logit adjustments $\boldsymbol{q}_c$ to determine unit contributions.
(c)~Specificity-regularized optimization promotes polarized shared gates, diverse conditional deviations, and appropriate deviation magnitudes, enabling shared and specialized contributions to emerge during pre-training.}
\label{fig:method}
\vspace{-0.1in}
\end{figure*}

\paragraph{Diffusion and Flow-based Generative Models.}
Diffusion and flow-based models generate samples by reversing a corruption process or transporting noise toward the data distribution. 
Let $u_0$ denote a clean sample, with $u_0=\mathcal{E}(x)$ for latent-space models using a pretrained encoder $\mathcal{E}$, and $u_0=x$ for pixel-space models. An intermediate state is defined as $u_t=a(t)u_0+b(t)\epsilon$, where $\epsilon\sim\mathcal{N}(0,I)$ and $a(t),b(t)$ specify the signal and noise coefficients.
Given a condition $c$, the network $f_\theta$ predicts a target $y_t$ using
\begin{equation}
\mathcal{L}_{\mathrm{gen}}
=
\mathbb{E}_{u_0,c,t,\epsilon}
\left[
    \left\|y_t-f_\theta(u_t,t,c)\right\|_2^2
\right].
\label{eq:generative_objective}
\end{equation}
where $y_t=\epsilon$ for noise prediction and $y_t=\dot{a}(t)u_0+\dot{b}(t)\epsilon$ for flow matching. Rectified flow is a special case with a linear interpolation path between data and noise.

\vspace{-0.1in}
\paragraph{Diffusion Transformers.}
A DiT processes patch tokens $X\in\mathbb{R}^{T\times D}$ through $L$ Transformer blocks, each containing multi-head self-attention (MSA) and a token-wise FFN. Each attention head computes $\operatorname{Attention}(Q,K,V)=\operatorname{softmax}(\frac{QK^\top}{\sqrt{d}})V$, where $Q,K,V$ are projected token features and $d$ is the head dimension. The head outputs are concatenated and projected back to dimension $D$.

The FFN independently transforms each token through
\begin{equation}
    \operatorname{FFN}_{\ell}(X)
    =
    \phi\!\left(XW_{\mathrm{in}}^\ell+b_{\mathrm{in}}^\ell\right)
    W_{\mathrm{out}}^\ell+b_{\mathrm{out}}^\ell,
    \label{eq:standard_ffn}
\end{equation}
where $W_{\mathrm{in}}^\ell\in\mathbb{R}^{D\times D'}$, $W_{\mathrm{out}}^\ell\in\mathbb{R}^{D'\times D}$, and $D'$ is the hidden dimension. adaLN-Zero conditions both residual branches on timestep and class embeddings.

\subsection{Specificity-Guided Knowledge Diversion}
\label{sec:method}
SGKDiv learns reusable components through adaptive routing and specialization regularization. We first introduce \textit{Native Paired Knowledge Units}  (Sec.~\ref{sec:paired_units}) to preserve FFN structure without SVD-based reparameterization, then learn their contributions through \textit{Calibrated Condition-Specific Routing} (Sec.~\ref{sec:calibrated_routing}). \textit{Specificity-Regularized Optimization} (Sec.~\ref{sec:specificity_regularization}) guides this allocation by regulating shared usage, conditional diversity, and specialization strength.

\subsubsection{Native Paired Knowledge Units}
\label{sec:paired_units}

Existing knowledge diversion methods~\cite{xie2025kind,xie2026knowledge,feng2025knowledge} construct Knowledge Units through matrix-wise SVD, introducing redundant factor parameters without explicitly coupling consecutive projections. 
SGKDiv instead defines Knowledge Units from paired native FFN channels, preserving this correspondence without factorization (see Fig.~\ref{fig:method}a).

Specifically, let $\mathbf{u}_j^\ell$ and $\mathbf{v}_j^{\ell\top}$ denote the $j$-th column of $W_{\mathrm{in}}^\ell$ and the corresponding row of $W_{\mathrm{out}}^\ell$, respectively. Omitting biases, the FFN in Eq.~\eqref{eq:standard_ffn} can be expressed as
\begin{equation}
\operatorname{FFN}_{\ell}(X)
=
\sum_{j=1}^{D'}
\phi\!\left(X\mathbf{u}_j^\ell\right)
\mathbf{v}_j^{\ell\top}.
\label{eq:channel_expansion}
\end{equation}
Each pair $\Theta_j^\ell=(\mathbf{u}_j^\ell,\mathbf{v}_j^\ell)$ defines a \textbf{\textit{Knowledge Unit}} spanning an input--output path, retaining the original parameters without factorization or reconstruction. 
Rather than preallocating Knowledge Units to condition-agnostic or condition-specific subsets before training, we learn their shared and specialized roles through routing during pre-training, allowing knowledge allocation to emerge from data.

\subsubsection{Calibrated Condition-Specific Routing}
\label{sec:calibrated_routing}

To learn knowledge allocation without a predefined partition, we combine a condition-agnostic gate $G_{\mathrm{agn}}^\ell$ with a condition-specific gate $G_{\mathrm{spe}}^\ell$ over all Knowledge Units, as shown in Fig.~\ref{fig:method}b. 
The former establishes a calibrated prior for shared unit usage, while the latter introduces condition-dependent adjustments. Together, they control each unit's contribution during pre-training.

Specifically, the condition-agnostic gate $G_{\mathrm{agn}}^\ell$ is parameterized by $\boldsymbol{s}^\ell\in\mathbb{R}^{D'}$, with one trainable parameter per Knowledge Unit. Its output is a shared usage prior $\boldsymbol{p}^\ell$:
\begin{equation}
    p_j^\ell
    =
    \operatorname{sigmoid}(s_j^\ell-\tau_\ell),
    \qquad
    \frac{1}{D'}\sum_{j=1}^{D'}p_j^\ell=\rho,
    \label{eq:calibration_constraint}
\end{equation}
where $\tau_\ell$ calibrates the mean gate value to $\rho$. The resulting prior controls unit contributions in the \textbf{\textit{unconditional branch}} without prescribing fixed shared-unit assignments.

Given a condition embedding $\mathbf{e}_c$, the condition-specific gate predicts $\boldsymbol{q}_c^\ell=G_{\mathrm{spe}}^\ell(\mathbf{e}_c)\in\mathbb{R}^{D'}$, modulating the shared prior in logit space:
\begin{equation}
    \boldsymbol{\sigma}_c^\ell
    =
    \operatorname{sigmoid}\!\left(
        \operatorname{logit}(\boldsymbol{p}^\ell)
        +\boldsymbol{q}_c^\ell
    \right).
    \label{eq:conditional_gate}
\end{equation}
The resulting coefficients modulate hidden channels before activation, yielding the condition-specific FFN:
\begin{equation}
\operatorname{CFFN}_{\ell}(X,c)
=
\phi\!\left(
    (XW_{\mathrm{in}}^\ell+b_{\mathrm{in}}^\ell)
    \odot\boldsymbol{\sigma}_c^\ell
\right)
\cdot W_{\mathrm{out}}^\ell+b_{\mathrm{out}}^\ell.
\label{eq:routed_ffn}
\end{equation}
Continuous condition-specific coefficients shape shared and specialized contributions during pre-training. For sparse inference, units with small coefficients are masked.

\subsubsection{Specificity-Regularized Optimization}
\label{sec:specificity_regularization}

Calibrated routing enables adaptive knowledge allocation, yet the generative objective $\mathcal{L}_{\mathrm{gen}}$ alone leaves specialization unconstrained. 
We therefore introduce \emph{Specificity-Regularized Optimization} to shape the shared usage prior and the diversity and magnitude of condition-specific deviations.

For a batch of $B$ conditional examples, we define condition-specific deviations from the shared prior as
\begin{equation}
    \boldsymbol{\delta}_i^\ell
    =
    \boldsymbol{\sigma}_{c_i}^\ell-\mathbf{p}^\ell
    \in\mathbb{R}^{D'},
    \label{eq:gate_residual}
\end{equation}
and stack them into $\boldsymbol{\Delta}^\ell=[\boldsymbol{\delta}_1^\ell,\ldots,\boldsymbol{\delta}_B^\ell]^\top\in\mathbb{R}^{B\times D'}$. 
We then apply \textit{gate polarization} to sharpen shared usage preferences and \textit{deviation diversity} and \textit{magnitude} losses to encourage distinct, appropriately scaled conditional adjustments.

\vspace{-0.1in}
\paragraph{Gate Polarization.}
The calibration constraint in Eq.~\eqref{eq:calibration_constraint} fixes the mean gate value at $\rho$ but permits intermediate coefficients. 
To encourage decisive shared usage preferences, as illustrated in Fig.~\ref{fig:method}c, we minimize
\begin{equation}
\mathcal{L}_{\mathrm{pol}}
=
\operatorname{Avg}_{\ell,j}
\left[p_j^\ell(1-p_j^\ell)\right],
\label{eq:polarization_loss}
\end{equation}
where $\operatorname{Avg}_{\ell,j}$ averages over routed layers and Knowledge Units. 
The loss promotes binary coefficients, while the fixed mean $\rho\in(0,1)$ prevents all-zero or all-one collapse.

\vspace{-0.1in}
\paragraph{Deviation Diversity.}
Condition-specific deviations should differ across conditions while allowing greater similarity between related conditions. Let $\boldsymbol{A}^\ell$ and $\boldsymbol{S}$ denote the pairwise cosine-similarity matrices of $\boldsymbol{\Delta}^\ell$ and the condition embeddings, respectively. 
We define an adaptive similarity threshold $\Gamma_{ij}=m+(1-m)[S_{ij}]_+$ and minimize
\begin{equation}
    \mathcal{L}_{\mathrm{div}}
    =
    \operatorname{Avg}_{\ell,\,i\not\sim j}
    \left[A_{ij}^\ell-\Gamma_{ij}\right]_+^2,
    \label{eq:dispersion_loss}
\end{equation}
where $[x]+=\max(0,x)$, $i\not\sim j$ denotes different-condition pairs, and $m\in[0,1]$ is the minimum threshold. Only similarities exceeding $\Gamma_{ij}$ are penalized, allowing greater alignment between conditions with similar embeddings.

\paragraph{Deviation Magnitude.}
Deviation diversity regulates directional differences but leaves their strength unconstrained.
We therefore center the deviations across the batch,
$\bar{\boldsymbol{\delta}}_i^\ell
=\boldsymbol{\delta}_i^\ell-\operatorname{Avg}_{k}\boldsymbol{\delta}_k^\ell$,
and regularize their root-mean-square (RMS) magnitude toward $\kappa>0$:
\begin{equation}
    \mathcal{L}_{\mathrm{mag}}
    =
    \operatorname{Avg}_{\ell,i}
    \left[
        \operatorname{RMS}(\bar{\boldsymbol{\delta}}_i^\ell)-\kappa
    \right]^2,
    \label{eq:magnitude_loss}
\end{equation}
where $\operatorname{RMS}(\mathbf{v})=\frac{\| \mathbf{v}\|_2}{\sqrt{D'}}$. Centering isolates variation across conditions, while RMS normalization makes $\kappa$ a per-channel measure of condition-specific variation.

The complete pre-training objective is
\begin{equation}
    \mathcal{L}
    =
    \mathcal{L}_{\mathrm{gen}}
    +\lambda_{\mathrm{pol}}\mathcal{L}_{\mathrm{pol}}
    +\alpha\left(
        \lambda_{\mathrm{div}}\mathcal{L}_{\mathrm{div}}
        +\lambda_{\mathrm{mag}}\mathcal{L}_{\mathrm{mag}}
    \right),
\label{eq:complete_objective}
\end{equation}
where $\alpha$ scales the diversity and magnitude penalties according to the uncertainty of the shared usage prior:
\begin{equation}
    \alpha
    =
    \operatorname{stopgrad}\!\left[
        \frac{\operatorname{Avg}_{\ell,j}[H(p_j^\ell)]}
             {H(\rho)}
    \right],
    \label{eq:uncertainty_annealing}
\end{equation}
with $H(u)=-u\log u-(1-u)\log(1-u)$. The fixed mean $\rho$ bounds the average entropy by $H(\rho)$, ensuring $\alpha\in[0,1]$. 
As the shared prior becomes polarized, these penalties weaken while agnostic polarization remains active.

\renewcommand{\arraystretch}{0.88}

\begin{table*}[t]
\centering
\caption{\textbf{Comparison with dense and knowledge diversion methods.}
We evaluate class-conditional generation on ImageNet-1K at $256\times256$ resolution across DiT, DiT-Flow, REPA, and JiT configurations. We report activated and total parameters (M), alongside FID$\downarrow$, IS$\uparrow$, Precision$\uparrow$, and Recall$\uparrow$ at CFG $1.0$ and $1.5$. All models are trained for 800K updates.}
\vspace{-0.1in}
\setlength{\tabcolsep}{5 pt}
\resizebox{\linewidth}{!}{
\begin{tabular}{@{}ll|cc|cccccccc@{}}
\toprule
\multirow{2}{*}{\textbf{Model}}
& \multirow{2}{*}{\textbf{Method}}
& \multirow{2}{*}{\#}\multirow{2}{*}{\shortstack[c]{\textbf{Activated} \\ \textbf{Param.}}}
& \multirow{2}{*}{\#}\multirow{2}{*}{\shortstack[c]{\textbf{Total} \\ \textbf{Param.}}}
& \multicolumn{4}{c}{\textbf{cfg=1.0}}
& \multicolumn{4}{c}{\textbf{cfg=1.5}} \\
\cmidrule(lr){5-8}
\cmidrule(lr){9-12}
& & & 
& \textbf{FID$\downarrow$}
& \textbf{IS$\uparrow$} 
& \textbf{Precision$\uparrow$} 
& \textbf{Recall$\uparrow$} 
& \textbf{FID$\downarrow$}
& \textbf{IS$\uparrow$} 
& \textbf{Precision$\uparrow$} 
& \textbf{Recall$\uparrow$} \\
\midrule
\multirow{4}{*}{\textbf{\texttt{DiT-S}}} 
& Baseline~\cite{peebles2023dit}
& 32.96 M & 32.96 M 
& 57.845 & 24.731 & 0.408 & 0.605 
& 33.645 & 47.460 & 0.538 & \textbf{0.567} \\
& KIND~\cite{xie2025kind}
& 32.49 M & 60.62 M 
& 57.340 & 25.425 & 0.412 & 0.599 
& 30.981 & 50.767 & 0.565 & 0.525 \\
& KDiv~\cite{xie2026knowledge}
& 31.17 M & 34.27 M 
& 55.912 & 26.020 & 0.424 & 0.595 
& 30.661 & 50.964 & 0.584 & 0.538 \\
& \cellcolor{blue!12}{SGKDiv}    
& \cellcolor{blue!12}{30.13 M} & \cellcolor{blue!12}{34.59 M} 
& \cellcolor{blue!12}{\textbf{54.092}} 
& \cellcolor{blue!12}{\textbf{28.231}} 
& \cellcolor{blue!12}{\textbf{0.425}} 
& \cellcolor{blue!12}{\textbf{0.606}}
& \cellcolor{blue!12}{\textbf{27.439}} 
& \cellcolor{blue!12}{\textbf{59.718}} 
& \cellcolor{blue!12}{\textbf{0.593}} 
& \cellcolor{blue!12}{0.526} \\
\addlinespace
\multirow{4}{*}{\textbf{\texttt{DiT-B}}} 
& Baseline~\cite{peebles2023dit}
& 130.51 M & 130.51 M
& 33.584 & 44.693 & 0.537 & 0.639
& 12.960 & 101.200 & 0.714 & \textbf{0.544} \\
& KIND~\cite{xie2025kind} 
& 128.91 M & 185.14 M
& 35.335 & 40.672 & 0.537 & 0.627 
& 14.525 & 87.567 & 0.712 & 0.528 \\
& KDiv~\cite{xie2026knowledge} 
& 123.09 M & 133.13 M 
& 32.704 & 46.512 & 0.539 & 0.631 
& 12.204 & 105.616 & 0.726 & 0.532 \\
& \cellcolor{blue!12}{SGKDiv}   
& \cellcolor{blue!12}{118.87 M} & \cellcolor{blue!12}{133.76 M}
& \cellcolor{blue!12}{\textbf{31.447}} 
& \cellcolor{blue!12}{\textbf{50.249}} 
& \cellcolor{blue!12}{\textbf{0.541}} 
& \cellcolor{blue!12}{\textbf{0.648}}
& \cellcolor{blue!12}{\textbf{10.030}} 
& \cellcolor{blue!12}{\textbf{122.589}} 
& \cellcolor{blue!12}{\textbf{0.733}} 
& \cellcolor{blue!12}{0.528}\\

\midrule
\multirow{2}{*}{\textbf{\texttt{DiT-Flow-S}}} 
& Baseline~\cite{liu2023rectifiedflow} 
& 32.96 M & 32.96 M
& 46.046 & 32.952 & 0.455 & 0.631
& 18.849 & 81.432 & 0.646 & \textbf{0.547} \\
& \cellcolor{blue!12}{SGKDiv}
& \cellcolor{blue!12}{30.39 M} & \cellcolor{blue!12}{34.58 M}
& \cellcolor{blue!12}{\textbf{44.031}} 
& \cellcolor{blue!12}{\textbf{35.902}} 
& \cellcolor{blue!12}{\textbf{0.462}} 
& \cellcolor{blue!12}{\textbf{0.632}}
& \cellcolor{blue!12}{\textbf{15.901}} 
& \cellcolor{blue!12}{\textbf{94.559}} 
& \cellcolor{blue!12}{\textbf{0.681}} 
& \cellcolor{blue!12}{0.516}\\
\addlinespace
\multirow{2}{*}{\textbf{\texttt{DiT-Flow-B}}}  
& Baseline~\cite{liu2023rectifiedflow} 
& 130.50 M & 130.50 M
& 26.228 & 58.394 & 0.562 & \textbf{0.663}
& 6.667 & 155.741 & 0.768 & \textbf{0.536} \\
& \cellcolor{blue!12}{SGKDiv}  
& \cellcolor{blue!12}{125.21 M} & \cellcolor{blue!12}{132.53 M}
& \cellcolor{blue!12}{\textbf{24.544}} 
& \cellcolor{blue!12}{\textbf{63.407}} 
& \cellcolor{blue!12}{\textbf{0.570}}
& \cellcolor{blue!12}{0.660}
& \cellcolor{blue!12}{\textbf{5.420}} 
& \cellcolor{blue!12}{\textbf{173.914}} 
& \cellcolor{blue!12}{\textbf{0.790}} 
& \cellcolor{blue!12}{0.522}\\
\midrule
\multirow{2}{*}{\textbf{\texttt{REPA-S}}} 
& Baseline~\cite{yu2025repa}   
& 39.52 M & 39.52 M 
& 42.541 & 35.474 & 0.482 & \textbf{0.634}
& 17.306 & 89.226 & 0.658 & \textbf{0.551} \\
& \cellcolor{blue!12}{SGKDiv}
& \cellcolor{blue!12}{37.07 M} & \cellcolor{blue!12}{41.14 M}
& \cellcolor{blue!12}{\textbf{40.587}} 
& \cellcolor{blue!12}{\textbf{39.733}} 
& \cellcolor{blue!12}{\textbf{0.489}} 
& \cellcolor{blue!12}{0.628}
& \cellcolor{blue!12}{\textbf{14.238}} 
& \cellcolor{blue!12}{\textbf{105.494}} 
& \cellcolor{blue!12}{\textbf{0.696}} 
& \cellcolor{blue!12}{0.515}\\
\addlinespace
\multirow{2}{*}{\textbf{\texttt{REPA-B}}} 
& Baseline~\cite{yu2025repa} 
& 137.85 M & 137.85 M 
& 17.917 & 80.206 & \textbf{0.621} & 0.658
& 3.810 & 209.129 & 0.806 & 0.535 \\
& \cellcolor{blue!12}{SGKDiv}   
& \cellcolor{blue!12}{132.69 M} & \cellcolor{blue!12}{139.68 M}
& \cellcolor{blue!12}{\textbf{17.884}} 
& \cellcolor{blue!12}{\textbf{81.799}} 
& \cellcolor{blue!12}{0.613} 
& \cellcolor{blue!12}{\textbf{0.661}} 
& \cellcolor{blue!12}{\textbf{3.623}} 
& \cellcolor{blue!12}{\textbf{217.165}} 
& \cellcolor{blue!12}{\textbf{0.809}} 
& \cellcolor{blue!12}{\textbf{0.544}} \\
\midrule
\multirow{2}{*}{\textbf{\texttt{JiT-S}}} 
& Baseline~\cite{li2026jit} 
& 33.42 M & 33.42 M 
& 80.170 & 17.145 & 0.280 & 0.520
& 62.302 & 25.390 & 0.345 & 0.540 \\
& \cellcolor{blue!12}{SGKDiv}  
& \cellcolor{blue!12}{33.04 M} & \cellcolor{blue!12}{34.63 M}
& \cellcolor{blue!12}{\textbf{76.257}} 
& \cellcolor{blue!12}{\textbf{18.541}} 
& \cellcolor{blue!12}{\textbf{0.289}} 
& \cellcolor{blue!12}{\textbf{0.535}}
& \cellcolor{blue!12}{\textbf{56.775}} 
& \cellcolor{blue!12}{\textbf{28.818}} 
& \cellcolor{blue!12}{\textbf{0.362}} 
& \cellcolor{blue!12}{\textbf{0.556}}\\
\addlinespace
\multirow{2}{*}{\textbf{\texttt{JiT-B}}} 
& Baseline~\cite{li2026jit} 
& 131.32 M & 131.32 M 
& 41.409 & 39.654 & \textbf{0.444} & 0.637
& 22.898 & 75.842 & 0.561 & \textbf{0.606} \\
& \cellcolor{blue!12}{SGKDiv}
& \cellcolor{blue!12}{129.48 M} & \cellcolor{blue!12}{135.37 M}
& \cellcolor{blue!12}{\textbf{40.613}}
& \cellcolor{blue!12}{\textbf{40.380}}
& \cellcolor{blue!12}{0.444} 
& \cellcolor{blue!12}{\textbf{0.649}}
& \cellcolor{blue!12}{\textbf{22.133}}
& \cellcolor{blue!12}{\textbf{76.666}}
& \cellcolor{blue!12}{\textbf{0.565}} 
& \cellcolor{blue!12}{0.606}\\
\bottomrule
\end{tabular}
}
\vspace{-0.1in}
\label{tab:main}
\end{table*}

\subsection{Adaptive Model Combination and Fine-Tuning}
\label{sec:adaptation}

For each unseen target dataset $\mathcal{D}_{\mathrm{tgt}}$, we optimize class embeddings for 300 steps with pre-trained parameters frozen. We then compute the mean gating coefficients in each FFN:
\begin{equation}
    \bar{\boldsymbol{\sigma}}^\ell
    =
    \frac{1}{|\mathcal{D}_{\mathrm{tgt}}|}
    \sum_{(x,c)\in\mathcal{D}_{\mathrm{tgt}}}
    \boldsymbol{\sigma}_c^\ell.
    \label{eq:target_relevance}
\end{equation}
We select the $K$ Knowledge Units with the highest mean coefficients in each layer:
\begin{equation}
    \mathcal{A}_\ell
    =
    \operatorname{TopKIndices}\!\left(
        \bar{\boldsymbol{\sigma}}^\ell,K
    \right).
    \label{eq:downstream_selection}
\end{equation}
Downstream fine-tuning updates only the selected units, yielding the trainable parameter set
\begin{equation}
    \Theta_{\mathrm{tgt}}
    =
    \bigcup_{\ell}
    \left\{\Theta_j^\ell \mid j\in\mathcal{A}_\ell\right\}
    =
    \bigcup_{\ell}
    \left\{
        (\mathbf{u}_j^\ell,\mathbf{v}_j^\ell)
        \mid j\in\mathcal{A}_\ell
    \right\},
\label{eq:adaptation_parameters}
\end{equation}
while the gating networks and all remaining backbone parameters remain frozen. 
This procedure translates the knowledge organization learned during pre-training into selective, target-aware updates, with $K$ controlling the adaptation budget.

\section{Experimental Setup}
\label{sec:experimental_setup}

\paragraph{Datasets and Evaluation.}
We pre-train class-conditional models on ImageNet-1K~\cite{deng2009imagenet} at $256\times256$ resolution and evaluate 50K generated samples using FID~\cite{heusel2017gans}, IS~\cite{salimans2016improved}, and Precision/Recall~\cite{kynkaanniemi2019improved}.
We assess downstream adaptation on eight datasets: CUB-200-2011~\cite{wah2011caltech}, DF20M~\cite{picek2022danish}, Flowers-102~\cite{nilsback2008automated}, Food-101~\cite{bossard2014food}, LSUN Bedroom and Church~\cite{yu2015lsun}, NWPU-RESISC45~\cite{cheng2017remote}, and Stanford Cars~\cite{krause20133d}. For each adapted model, we generate 10K samples at CFG $1.5$ and report FD-DINOv2, KD-DINOv2~\cite{stein2023exposing}, and CMMD~\cite{jayasumana2024rethinking}.

\vspace{-0.1in}
\paragraph{Basic Training Settings.}
We pre-train models for 800K updates with a global batch size of $256$ unless otherwise specified.
Latent-space models use patch size $2$ and AdamW with learning rate $10^{-4}$ and zero weight decay.
We use an EMA decay of $0.9999$ and classifier-free label dropout of $0.1$.
REPA employs a frozen DINOv2-ViT-B encoder for representation alignment.
Pixel-space JiT uses five warm-up epochs and a base learning rate of $5\times10^{-5}$.
For the MoE comparison, we follow the ProMoE training budget of 500K updates.

\section{Results}

\begin{table*}[t]
\centering
\caption{\textbf{Comparison with mixture-of-experts methods.}
We compare SGKDiv with DiTMoE, EC-DiT, and ProMoE variants using expert-choice (EC) and token-choice (TC) routing on ImageNet-1K at $256\times256$ resolution. We report activated and total parameters (M), alongside FID$\downarrow$, IS$\uparrow$, Precision$\uparrow$, and Recall$\uparrow$ at CFG $1.0$ and $1.5$. All models are trained for 500K updates following the ProMoE protocol.}
\vspace{-0.1in}
\setlength{\tabcolsep}{6 pt}
\resizebox{\linewidth}{!}{
\begin{tabular}{@{}ll|cc|cccccccc@{}}
\toprule
\multirow{2}{*}{\textbf{Model}}
& \multirow{2}{*}{\textbf{Method}}
& \multirow{2}{*}{\#}\multirow{2}{*}{\shortstack[c]{\textbf{Activated} \\ \textbf{Param.}}}
& \multirow{2}{*}{\#}\multirow{2}{*}{\shortstack[c]{\textbf{Total} \\ \textbf{Param.}}}
& \multicolumn{4}{c}{\textbf{cfg=1.0}}
& \multicolumn{4}{c}{\textbf{cfg=1.5}} \\
\cmidrule(lr){5-8}
\cmidrule(lr){9-12}
& & & 
& \textbf{FID$\downarrow$}
& \textbf{IS$\uparrow$} 
& \textbf{Precision$\uparrow$} 
& \textbf{Recall$\uparrow$} 
& \textbf{FID$\downarrow$}
& \textbf{IS$\uparrow$} 
& \textbf{Precision$\uparrow$} 
& \textbf{Recall$\uparrow$} \\
\midrule
\multirow{5}{*}{\textbf{\texttt{DiT-S}}} 
& DiTMoE~\cite{fei2024scaling}
& 32.98 M & 82.61 M 
& 61.339 & 22.427 & \textbf{0.398} & 0.569 
& 38.652 & 39.136 & 0.512 & 0.547 \\
& EC-DiT~\cite{sun2025ecdit}
& 32.98 M & 82.61 M 
& 62.967 & 22.212 & 0.390 & 0.568 
& 39.541 & 38.993 & 0.505 & 0.552 \\
& ProMoE-EC~\cite{wei2026routing}
& 32.99 M & 75.54 M
& 64.320 & 21.885 & 0.376 & 0.571 
& 40.184 & 39.155 & 0.493 & 0.560 \\
& ProMoE-TC~\cite{wei2026routing}
& 32.99 M & 75.54 M 
& 62.619 & 22.832 & 0.387 & 0.581 
& 38.147 & 41.707 & 0.506 & \textbf{0.580} \\
& \cellcolor{blue!12}{SGKDiv}    
& \cellcolor{blue!12}{31.79 M} & \cellcolor{blue!12}{34.59 M} 
& \cellcolor{blue!12}{\textbf{59.783}} 
& \cellcolor{blue!12}{\textbf{24.598}} 
& \cellcolor{blue!12}{0.395} 
& \cellcolor{blue!12}{\textbf{0.584}}
& \cellcolor{blue!12}{\textbf{33.070}} 
& \cellcolor{blue!12}{\textbf{48.850}} 
& \cellcolor{blue!12}{\textbf{0.546}} 
& \cellcolor{blue!12}{0.536} \\
\midrule
\multirow{5}{*}{\textbf{\texttt{DiT-B}}} 
& DiTMoE~\cite{fei2024scaling}
& 130.55 M & 328.89M 
& 39.595 & 37.712 & 0.500 & 0.636 
& 17.828 & 80.393 & 0.662 & 0.551 \\
& EC-DiT~\cite{sun2025ecdit}
& 130.55 M  & 328.89 M
& 37.796 & 38.874 & \textbf{0.512} & 0.623 
& 16.750 & 82.827 & 0.677 & 0.551 \\
& ProMoE-EC~\cite{wei2026routing}
& 130.57 M & 300.61 M 
& 41.635 & 35.849 & 0.491 & 0.632 
& 19.030 & 77.735 & 0.642 & \textbf{0.564} \\
& ProMoE-TC~\cite{wei2026routing}
& 130.57 M & 300.61 M
& 39.445 & 38.300 & 0.502 & 0.633 
& 17.351 & 83.629 & 0.659 & 0.564 \\
& \cellcolor{blue!12}{SGKDiv}   
& \cellcolor{blue!12}{118.61 M} & \cellcolor{blue!12}{133.76 M} 
& \cellcolor{blue!12}{\textbf{36.424}} 
& \cellcolor{blue!12}{\textbf{42.498}} 
& \cellcolor{blue!12}{0.511} 
& \cellcolor{blue!12}{\textbf{0.638}}
& \cellcolor{blue!12}{\textbf{13.482}} 
& \cellcolor{blue!12}{\textbf{101.290}} 
& \cellcolor{blue!12}{\textbf{0.699}} 
& \cellcolor{blue!12}{0.532}\\
\bottomrule
\end{tabular}
}
\vspace{-0.08in}
\label{tab:moe}
\end{table*}

\begin{figure*}[tb]
\centering
\includegraphics[width=0.96\linewidth]{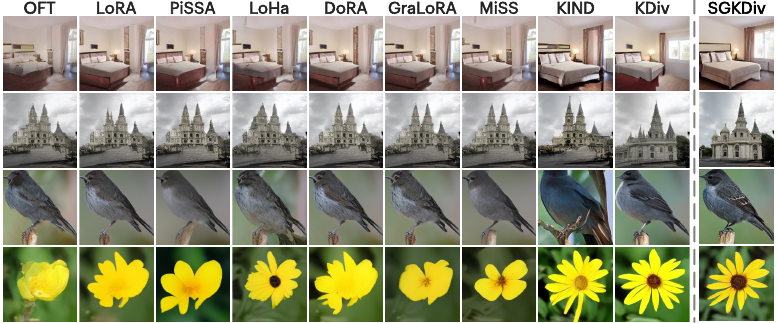}
\vspace{-0.1in}
\caption{\textbf{Qualitative comparison of downstream adaptation.}
Samples from novel target classes generated by SGKDiv and competing methods at $256\times256$ resolution with CFG $1.5$.}
\vspace{-0.14in}
\label{fig:downstream}
\end{figure*}

\subsection{Performance on the Pre-Training Distribution}
\label{sec:in_distribution}
\paragraph{Comparison with Dense and Knowledge Diversion Methods.}
Table~\ref{tab:main} compares SGKDiv with dense and knowledge diversion baselines, showing consistent FID and IS improvements. At CFG $1.5$, SGKDiv reduces FID by $6.206$ and $2.930$ on DiT-S and DiT-B over dense baselines, and by $3.222$ and $2.174$ over KDiv, respectively. Gains extend to flow matching (i.e., DiT-Flow), representation alignment (i.e., REPA), and pixel-space generation (i.e., JiT), demonstrating applicability across training formulations and representation spaces while retaining native objectives.

Compared with dense models, SGKDiv introduces a condition-aware inductive bias through regularized routing that balances shared computation and specialization. On DiT-B, it improves generation quality while reducing activated parameters from $130.51$M to $118.87$M, with total parameters increasing modestly to $133.76$M. These results suggest that condition-dependent channel allocation enables more effective use of activated capacity.

Compared with previous knowledge diversion methods, SGKDiv offers a more compact and flexible parameterization. Native channel pairing avoids SVD-related overhead, reducing DiT-S total parameters from KIND's $60.62$M to $34.59$M. Against KDiv, SGKDiv improves FID, IS, and Precision with comparable total parameters and fewer activated parameters, indicating benefits beyond compactness. These gains are consistent with adaptive routing and explicit regularization enabling more effective specialization than fixed shared--specific assignments alone.

\vspace{-0.15in}
\paragraph{Comparison with MoE Baselines.}
Table~\ref{tab:moe} compares SGKDiv with representative MoE baselines. On DiT-S, SGKDiv achieves the lowest FID and highest IS at both guidance scales. 
At CFG $1.5$, it lowers FID from ProMoE-TC's $38.147$ to $33.070$, while reducing total parameters from $75.54$M to $34.59$M and activated parameters from $32.99$M to $31.79$M. These results demonstrate stronger generation performance with a substantially more compact model.

Unlike token-wise expert allocation, SGKDiv modulates native paired channels according to conditions and explicitly regularizes their specialization.
This provides a consistent condition-level allocation across tokens, allowing differentiated computation without replicating full FFN experts. 
The gains at comparable activated parameter counts and substantially lower total capacity suggest that effective specialization benefits from parameter organization and training beyond expert expansion, positioning SGKDiv as a compact alternative for improving generation quality.

\begin{table*}[t]
\centering
\setlength{\tabcolsep}{0.8 mm}
\caption{\textbf{Performance of SGKDiv in downstream adaptation.}
We compare SGKDiv with PEFT and knowledge diversion methods on datasets using FDD$\downarrow$, KDD$\downarrow$, and CMMD$\downarrow$. ``GPU Time'' and ``Param.'' denote adaptation time and trainable parameter count, respectively.}
\vspace{-0.1in}
\resizebox{\linewidth}{!}{
\begin{tabular}{@{}lcccccccccccc|ccc@{}}
\toprule[1pt]
& \multicolumn{3}{c}{\textsc{CUB-200}} & \multicolumn{3}{c}{\textsc{DF20M}} & \multicolumn{3}{c}{\textsc{Flowers-102}} & \multicolumn{3}{c|}{\textsc{Food-101}} & \multicolumn{3}{c}{\textsc{Cost}}\\ 
\cmidrule(lr){2-4} 
\cmidrule(lr){5-7}
\cmidrule(lr){8-10}
\cmidrule(lr){11-13} 
\cmidrule(l){14-16} 
& FDD$\downarrow$ & KDD$\downarrow$ & CMMD$\downarrow$ & FDD$\downarrow$ & KDD$\downarrow$ & CMMD$\downarrow$ & FDD$\downarrow$ & KDD$\downarrow$ & CMMD$\downarrow$ & FDD$\downarrow$ & KDD$\downarrow$ & CMMD$\downarrow$ & \multicolumn{2}{c}{\textsc{GPU Time}} & \textsc{Param.} \\ 
\cmidrule(r){1-13}
\cmidrule(l){14-16}
OFT~\cite{qiu2023controlling}
& 1130 & 1.452 & 0.635
& 977 & 3.213 & 1.366
& 1651 & 1.992 & 1.175
& 1290 & 2.979 & 2.217
& \multicolumn{2}{c}{1.26 h} & 15.03 M \\
LoRA~\cite{hulora}
& 1095 & 1.313 & 0.535
& 1053 & 3.544 & 1.387
& 1575 & 1.674 & 0.889
& 1245 & 2.835 & 2.007
& \multicolumn{2}{c}{1.13 h} & 11.80 M \\
PiSSA~\cite{meng2024pissa}
& 1051 & 1.230 & 0.531
& 930 & 3.089 & 1.263
& 1596 & 1.767 & 0.973
& 1208 & 2.697 & 1.964
& \multicolumn{2}{c}{1.14 h} & 11.80 M \\
LoHa~\cite{yeh2024navigating}
& 1313 & 1.819 & 0.704
& 1262 & 4.274 & 1.663
& 1649 & 1.920 & 0.877
& 1349 & 3.107 & 2.216
& \multicolumn{2}{c}{1.85 h} & 11.80 M \\
DoRA~\cite{liudora}
& 1077 & 1.286 & 0.557
& 1037 & 3.516 & 1.377
& 1553 & 1.652 & 0.959
& 1245 & 2.865 & 1.992
& \multicolumn{2}{c}{1.38 h} & 11.82 M \\
GraLoRA~\cite{jung2026gralora}
& 1156 & 1.443 & 0.569
& 1109 & 3.761 & 1.454
& 1581 & 1.717 & 0.848
& 1279 & 2.922 & 2.069
& \multicolumn{2}{c}{1.38 h} & 11.80 M \\
MiSS~\cite{kang2026miss}
& 1094 & 1.312 & 0.538
& 990 & 3.321 & 1.322
& 1574 & 1.665 & 0.924
& 1225 & 2.749 & 1.997
& \multicolumn{2}{c}{1.10 h} & 11.80 M \\
\cmidrule(r){1-13} 
\cmidrule(l){14-16} 
KIND~\cite{xie2025kind}
& 1053 & 1.246 & 0.568
& 803 & 2.682 & 1.120
& \textbf{1421} & \textbf{1.453} & 0.683
& 1177 & 2.655 & 1.817
& \multicolumn{2}{c}{1.22 h} & 11.80 M \\
KDiv~\cite{xie2026knowledge}
& 995 & 1.098 & 0.386
& 698 & 2.306 & 0.863
& 1507 & 1.576 & 0.685
& 1075 & 2.341 & 1.568
& \multicolumn{2}{c}{0.90 h} & 11.80 M \\
\midrule
\cellcolor{blue!10}{SGKDiv}
& \cellcolor{blue!10}{\textbf{945}} & \cellcolor{blue!10}{\textbf{1.016}} & \cellcolor{blue!10}{\textbf{0.356}}
& \cellcolor{blue!10}{\textbf{609}} & \cellcolor{blue!10}{\textbf{1.991}} & \cellcolor{blue!10}{\textbf{0.813}}
& \cellcolor{blue!10}{1473} & \cellcolor{blue!10}{1.457} & \cellcolor{blue!10}{\textbf{0.659}}
& \cellcolor{blue!10}{\textbf{1011}} & \cellcolor{blue!10}{\textbf{2.186}} & \cellcolor{blue!10}{\textbf{1.447}}
& \multicolumn{2}{c}{\cellcolor{blue!10}{0.90 h}} & \cellcolor{blue!10}{11.80 M} \\
\midrule[1pt]

& \multicolumn{3}{c}{\textsc{LSUN Bedroom}} & \multicolumn{3}{c}{\textsc{LSUN Church}} & \multicolumn{3}{c}{\textsc{RESISC45}} & \multicolumn{3}{c|}{\textsc{Stanford Cars}} & \multicolumn{3}{c}{\textsc{Average}} \\ 
\cmidrule(lr){2-4} 
\cmidrule(lr){5-7}
\cmidrule(lr){8-10}
\cmidrule(lr){11-13} 
\cmidrule(l){14-16} 
& FDD$\downarrow$ & KDD$\downarrow$ & CMMD$\downarrow$ & FDD$\downarrow$ & KDD$\downarrow$ & CMMD$\downarrow$ & FDD$\downarrow$ & KDD$\downarrow$ & CMMD$\downarrow$ & FDD$\downarrow$ & KDD$\downarrow$ & CMMD$\downarrow$ & FDD$\downarrow$ & KDD$\downarrow$ & CMMD$\downarrow$\\ 
\cmidrule(r){1-13}
\cmidrule(l){14-16}
OFT~\cite{qiu2023controlling}
& 1282 & 4.064 & 2.389
& 1304 & 3.116 & 2.598
& 913 & 3.573 & 1.711
& 730 & 2.029 & 0.918
& 1160 & 2.802 & 1.626 \\
LoRA~\cite{hulora}
& 1285 & 4.035 & 2.287
& 1301 & 3.129 & 2.605
& 703 & 2.962 & 1.503
& 692 & 1.755 & 0.899
& 1119 & 2.656 & 1.514 \\
PiSSA~\cite{meng2024pissa}
& 1202 & 3.806 & 2.181
& 1241 & 2.908 & 2.468
& 658 & 2.611 & 1.465
& 615 & 1.535 & 0.796
& 1063 & 2.455 & 1.455 \\
LoHa~\cite{yeh2024navigating}
& 1496 & 4.559 & 2.574
& 1406 & 3.479 & 2.854
& 826 & 3.485 & 1.598
& 908 & 2.514 & 1.275
& 1276 & 3.145 & 1.720 \\
DoRA~\cite{liudora}
& 1279 & 4.045 & 2.290
& 1315 & 3.193 & 2.616
& 699 & 2.940 & 1.499
& 708 & 1.896 & 0.871
& 1114 & 2.674 & 1.520 \\
GraLoRA~\cite{jung2026gralora}
& 1346 & 4.190 & 2.361
& 1335 & 3.256 & 2.678
& 736 & 3.106 & 1.531
& 759 & 1.987 & 0.988
& 1163 & 2.798 & 1.563 \\
MiSS~\cite{kang2026miss}
& 1256 & 3.937 & 2.233
& 1270 & 3.013 & 2.509
& 687 & 2.767 & 1.490
& 672 & 1.712 & 0.850
& 1096 & 2.560 & 1.483 \\
\cmidrule(r){1-13} 
\cmidrule(l){14-16} 
KIND~\cite{xie2025kind}
& 1101 & 3.660 & 1.959
& 1190 & 2.897 & 2.236
& \textbf{566} & \textbf{2.319} & \textbf{1.384}
& \textbf{554} & \textbf{1.386} & 0.622
& 983 & 2.287 & 1.299 \\
KDiv~\cite{xie2026knowledge}
& 1052 & 3.616 & 1.881
& 1130 & 2.701 & 2.057
& 665 & 2.746 & 1.470
& 604 & 1.530 & 0.608
& 966 & 2.239 & 1.190 \\
\midrule
\cellcolor{blue!10}{SGKDiv}
& \cellcolor{blue!10}{\textbf{962}} & \cellcolor{blue!10}{\textbf{3.380}} & \cellcolor{blue!10}{\textbf{1.769}}
& \cellcolor{blue!10}{\textbf{1093}} & \cellcolor{blue!10}{\textbf{2.580}} & \cellcolor{blue!10}{\textbf{1.935}}
& \cellcolor{blue!10}{619} & \cellcolor{blue!10}{2.491} & \cellcolor{blue!10}{1.441}
& \cellcolor{blue!10}{566} & \cellcolor{blue!10}{1.405} & \cellcolor{blue!10}{\textbf{0.547}}
& \cellcolor{blue!10}{\textbf{910}} & \cellcolor{blue!10}{\textbf{2.063}} & \cellcolor{blue!10}{\textbf{1.121}} \\
\bottomrule[1pt]
\end{tabular}
}
\vspace{-0.15in}
\label{tab:novel}
\end{table*}

\subsection{Performance in Downstream Adaptation}
\label{sec:downstream}
Table~\ref{tab:novel} evaluates downstream adaptation on eight datasets beyond the pre-training distribution. SGKDiv achieves the lowest average FDD, KDD, and CMMD of $910$, $2.063$, and $1.121$, respectively, outperforming both PEFT and knowledge diversion baselines overall.
Gains across fine-grained categories, natural scenes, and remote-sensing imagery support effective component reuse under diverse domain shifts. Representative generated samples are shown in Fig.~\ref{fig:downstream}.

Compared with PEFT methods, SGKDiv uses pre-trained routing to concentrate updates on target-relevant Knowledge Units. Under the same trainable parameter budget as PiSSA, it reduces average FDD from $1063$ to $910$ and adaptation time from $1.14$ to $0.90$ GPU hours. These gains suggest that target-aware component selection improves the effectiveness of limited updates by aligning pre-trained knowledge organization with downstream requirements.

The gains over existing knowledge diversion methods suggest that effective transfer depends on how components are learned, beyond structural separation. 
Fixed assignments constrain knowledge allocation around source conditions, which may not match downstream requirements. SGKDiv instead learns continuous contributions over native paired channels, with regularization promoting differentiated conditional adjustments without rigid component boundaries. Target-aware selection identifies useful units for joint adaptation while preserving unselected parameters, connecting pre-trained specialization to new-domain requirements and supporting stronger aggregate adaptation performance.

\begin{figure}[tb]
\centering
\includegraphics[width=\linewidth]{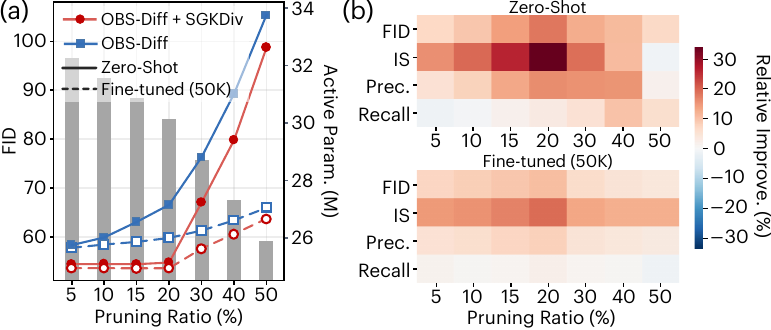}
\vspace{-0.28in}
\caption{\textbf{Compatibility of SGKDiv with model pruning.}
(a)~FID of OBS-Diff with and without SGKDiv and (b)~corresponding relative improvements across pruning ratios, evaluated before and after 50K fine-tuning steps.}
\vspace{-0.12in}
\label{fig:prune}
\end{figure}

\begin{figure*}[!t]
\centering
\includegraphics[width=0.98\linewidth]{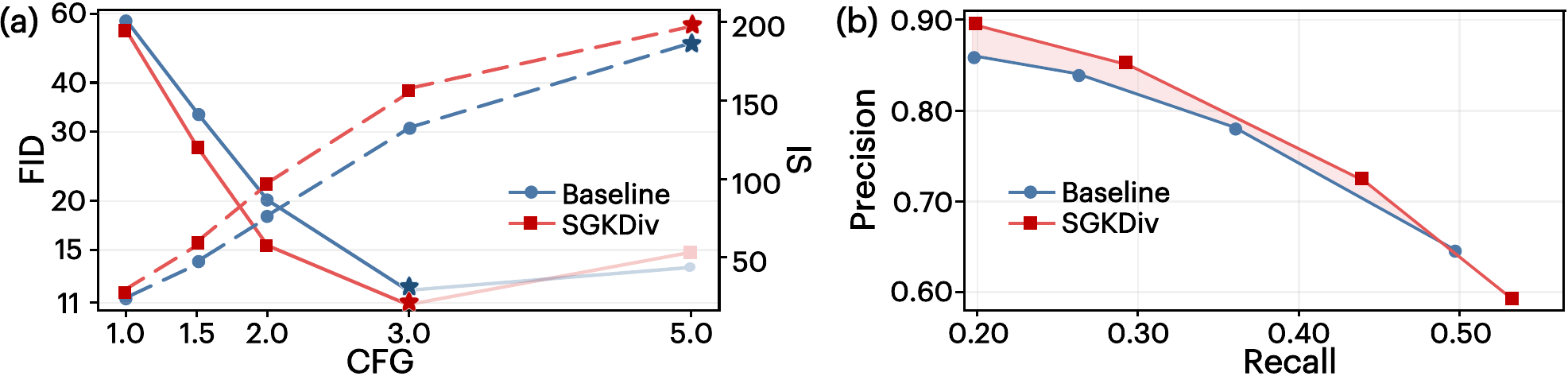}
\vspace{-0.12in}
\caption{\textbf{Generation performance under varying classifier-free guidance (CFG).}
(a)~FID and IS across CFG scales. (b)~Precision--Recall trade-offs across the same CFG scales. SGKDiv improves quality over a broad guidance range while balancing fidelity and coverage.}
\vspace{-0.12in}
\label{fig:cfg}
\end{figure*}

\begin{figure*}[!t]
\centering
\includegraphics[width=0.98\linewidth]{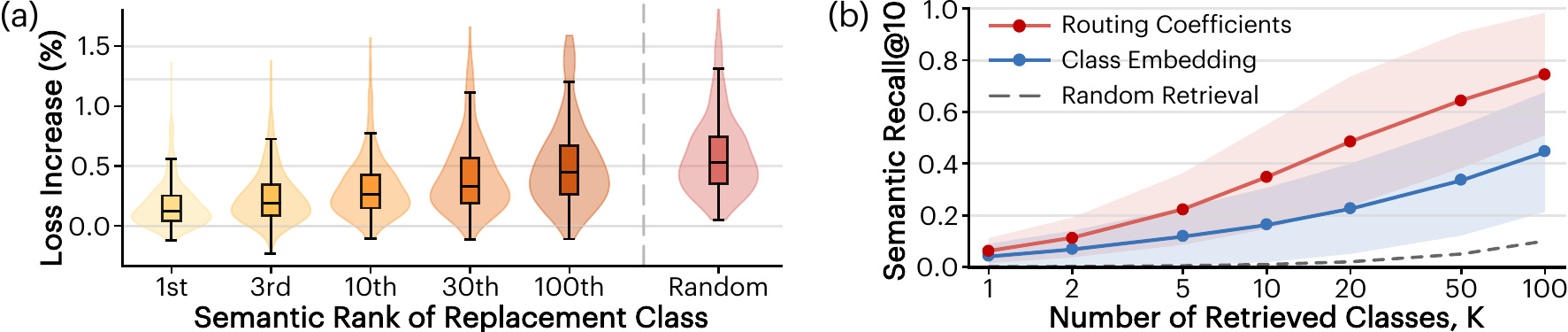}
\vspace{-0.1in}
\caption{\textbf{Semantic structure of condition-specific knowledge allocation.}
(a)~Loss increase under router-embedding replacements at different CLIP-based semantic ranks and random replacement. Semantically closer replacements incur smaller increases, indicating structured cross-condition reuse. 
(b)~Recall of CLIP top-10 neighbors among the $K$ nearest classes by routing-coefficient similarity.}
\vspace{-0.12in}
\label{fig:semantic}
\end{figure*}

\subsection{Performance Under Model Compression}
We combine SGKDiv with OBS-Diff~\cite{zhu2026obs} to assess whether learned knowledge organization supports model pruning. 
Fig.~\ref{fig:prune} shows that SGKDiv maintains nearly constant zero-shot FID up to $20\%$ pruning, whereas OBS-Diff deteriorates progressively. This stability indicates greater tolerance to moderate capacity reduction, consistent with learned allocation helping retained parameters preserve useful shared and conditional computation. The performance advantage also persists after 50K fine-tuning steps. Although aggressive pruning degrades both methods, SGKDiv consistently retains better generation quality, supporting the compatibility of its learned knowledge organization with model compression.

\subsection{Ablation and Analysis}
\begin{table}[t]
\centering
\setlength{\tabcolsep}{2 mm}
\caption{\textbf{Ablation of routing and specialization regularization.}
We evaluate the learnable condition-agnostic gate, regularization losses, and polarization-guided annealing. ``w/o Annealing'' fixes $\alpha=1$, disabling annealing of the diversity and magnitude losses.}
\vspace{-0.1in}
\resizebox{\linewidth}{!}{
\begin{tabular}{@{}lcccc@{}}
    \toprule[1pt]
    Method & FID$\downarrow$ & IS$\uparrow$ & Precision$\uparrow$ & Recall$\uparrow$ \\
    \midrule
    \rowcolor{blue!10}
    SGKDiv & \textbf{54.092} & \textbf{28.231} & \textbf{0.425} & 0.606 \\
    \midrule
    w/o Annealing $\alpha$ & 56.696 & 26.583 & 0.409 & \textbf{0.607} \\
    w/o $\mathcal{L}_{\mathrm{pol}}$ & 57.618 & 24.862 & 0.413 & 0.604 \\
    w/o $\mathcal{L}_{\mathrm{div}}$ & 58.514 & 24.314 & 0.410 & 0.604 \\
    w/o $\mathcal{L}_{\mathrm{mag}}$ & 55.516 & 27.138 & 0.419 & 0.605 \\
    \midrule
    w/o Learnable $G_{\mathrm{agn}}$ & 58.492 & 24.471 & 0.408 & 0.596 \\
   \bottomrule[1pt]
\end{tabular}
}
\vspace{-0.08in}
\label{tab:ablation}
\end{table}

\subsubsection{Ablation on Specificity-Regularized Optimization}

Table~\ref{tab:ablation} shows that removing any loss degrades generation performance. The largest deterioration occurs without $\mathcal{L}_{\mathrm{div}}$, which increases FID from $54.092$ to $58.514$. These losses address complementary aspects of knowledge allocation, with $\mathcal{L}_{\mathrm{pol}}$ sharpening shared usage preferences and $\mathcal{L}_{\mathrm{div}}$ and $\mathcal{L}_{\mathrm{mag}}$ regulating the diversity and strength of conditional adjustments. The results support explicitly guiding both shared allocation and conditional specialization.

Disabling polarization-guided annealing increases FID to $56.696$ despite retaining all losses, suggesting that fixed regularization can constrain later-stage optimization. Annealing relaxes the diversity and magnitude penalties as the shared prior polarizes, balancing early specialization guidance with subsequent generative refinement.

\subsubsection{Ablation on Learnable Condition-Agnostic Priors}

Table~\ref{tab:ablation} shows that fixing the condition-agnostic prior (i.e., w/o Learnable $G_{\mathrm{agn}}$) increases FID from $54.092$ to $58.492$, highlighting the value of learning shared allocation beyond controlling mean gate capacity. Independent of the input condition, this prior integrates training signals across conditions to establish a common reference for conditional specialization. 
A fixed prior preserves aggregate capacity but cannot adapt shared allocation during pre-training, potentially forcing conditional adjustments to compensate for an unsuitable reference. 
The results suggest that effective knowledge diversion requires both a learned condition-agnostic foundation and differentiated conditional contributions.

\subsubsection{Analysis of Performance across CFG Scales}
We compare SGKDiv with the dense baseline across CFG scales. Fig.~\ref{fig:cfg} shows that both attain their lowest FID near CFG $3.0$, with SGKDiv achieving a lower minimum FID and consistently higher IS. The Precision--Recall curves show higher precision at comparable low-to-moderate recall, with diminishing gains as coverage increases, suggesting improved fidelity rather than uniformly expanded coverage.
Since CFG amplifies differences between conditional and unconditional predictions, the learned shared prior and regularized conditional adjustments may support more effective guidance. Gains across CFG scales suggest that this benefit extends beyond a single sampling configuration.

\subsubsection{Analysis of Semantic Knowledge Organization}
We assess the semantic organization of learned routing using CLIP text embeddings~\cite{radford2021learning} as an external reference.
In Fig.~\ref{fig:semantic}(a), replacing only the router embedding with that of a more distant class yields a larger loss increase.
This suggests that conditional contributions capture semantic distinctions while preserving computation reusable across related conditions.
Fig.~\ref{fig:semantic}(b) shows that neighbors retrieved from the full routing coefficients recover CLIP top-10 neighbors above a random-retrieval baseline.
Together, these results suggest that specialization preserves cross-condition semantic relationships, supporting selective downstream reuse.

\section{Conclusion}
We presented \textit{Specificity-Guided Knowledge Diversion} (SGKDiv), which combines native FFN channel pairing, calibrated routing, and specialization regularization to learn reusable knowledge organization without SVD-based reparameterization or predefined shared--specific partitions. The resulting routing guides target-aware unit selection for downstream fine-tuning. 
Experiments demonstrate improved in-distribution generation over dense and MoE baselines and stronger downstream adaptation than representative PEFT methods under comparable trainable parameter budgets.

\section*{LLM Usage Statement}

The authors used generative AI tools, specifically GPT and Codex powered by GPT models, to assist with writing, LaTeX and figure formatting, and code editing. All claims, citations, code, figures, and numerical results were independently reviewed and verified by the authors, who take full responsibility for the final submission.

{
    \small
    \bibliographystyle{ieeenat_fullname}
    \bibliography{main}
}


\end{document}